\documentclass[draft]{agujournal2019}
\usepackage{url} 
\usepackage{lineno}
\usepackage{amsmath}
\usepackage{amsfonts}
\usepackage{bm}
\usepackage{float}
\usepackage[inline]{trackchanges} 
\usepackage{soul}
\usepackage{array}
\usepackage{caption}
\newcolumntype{P}[1]{>{\raggedright\arraybackslash}p{#1}}
\draftfalse

\journalname{Geophysical Research Letters}

\begin{document}
\nocite{*}

%
%


\title{Attention-Based Reconstruction of Full-Field Tsunami Waves from Sparse Tsunameter Networks}

%
%




\authors{Edward McDugald\affil{1,2}, Arvind Mohan\affil{2}, Darren Engwirda\affil{3,5}, Agnese Marcato\affil{4}, Javier Santos\affil{4}}

\affiliation{1}{University of Arizona}
\affiliation{2}{Computer, Computational, and Statistical Sciences Division, Los Alamos National Laboratory}
\affiliation{3}{Theoretical Division, Los Alamos National Laboratory}
\affiliation{4}{Earth and Environmental Sciences, Los Alamos National
Laboratory}
\affiliation{5}{Environment, Commonwealth Scientific \& Industrial Research Organisation}




\correspondingauthor{Edward McDugald}{=emcdugald@arizona.edu=}



\begin{keypoints}
\item An attention-based neural network specialized for sparse sensing is implemented to reconstruct tsunami waves from sparse observations.
\item We test our model on tsunami simulations whose epicenters differ from those used in training, replicating realistic forecasting conditions.
\item We compare our method to the Linear Interpolation with Huygens-Fresnel Principle to generate virtual waveforms, showing improved accuracy.
\end{keypoints}

%
%

%
%


\begin{abstract}
We investigate the potential of an attention-based neural network architecture, the Senseiver, for sparse sensing in tsunami forecasting. Specifically, we focus on the Tsunami Data Assimilation Method, which generates forecasts from tsunameter networks. Our model is used to reconstruct high-resolution tsunami wavefields from extremely sparse observations, including cases where the tsunami epicenters are not represented in the training set. Furthermore, we demonstrate that our approach significantly outperforms the Linear Interpolation with Huygens-Fresnel Principle in generating dense observation networks, achieving markedly improved accuracy.
\end{abstract}

\section*{Plain Language Summary}

While machine learning methods have achieved accurate forecasts of tsunami waveforms at fixed observation points, the full-field reconstruction of tsunami waves from sparse observations has not yet been demonstrated using machine learning techniques. This challenge is highly relevant to tsunami data assimilation, where in-ocean tsunameter observations are integrated with numerical models to improve tsunami forecasts. In this study, we employ an attention-based neural network architecture, the Senseiver, to produce high-fidelity reconstructions of tsunami wavefields from extremely sparse tsunameter measurements at realistic sensor locations. To illustrate the practical value of our approach for tsunami forecasters, we perform experiments in which the Senseiver generates virtual waveforms within a sparse observation network, and we compare its performance to a widely used interpolation method for the same task.

%
%

\section{Introduction}

Tsunami early warning systems are critical for mitigating the devastating human and economic impacts of these natural hazards. Most operational systems rely on partial differential equation (PDE)-based solvers, where the initial ocean surface displacement is inferred from earthquake source parameters and then propagated using shallow water equation (SWE) models. Widely adopted frameworks include NOAA’s MOST system \cite{Titov2016,titov1997implementation}, GeoClaw \cite{berger2011geoclaw,leveque2011tsunami}, and Gerris/Basilisk \cite{popinet2012adaptive,popinet2020vertically}. Despite broad use, these approaches face significant challenges, including uncertainty in seismic parameter estimation, high computational demand, and limited real-time adaptability.

To overcome these limitations and enable rapid response, tsunami forecasting increasingly leverages sparse offshore and onshore observations for source characterization and direct wave prediction. Methods such as DART buoy data inversion \cite{percival2011}, direct energy estimation from deep-ocean pressure measurements \cite{bernard2013}, and real-time data assimilation using GNSS and other sensors \cite{Hiroaki2014} have improved speed and accuracy, yet perform best when observation networks are sufficiently dense and well-placed. This has motivated the development of sparse sensing strategies to enhance observational coverage through techniques such as the generation of virtual waveform observations and optimization of sensor placement. For example, Wang et al.'s \cite{wang2019} Huygens-Fresnel interpolation synthesizes virtual tsunami waveforms from sparse tsunameter observations, effectively constructing a denser network that can then be used as input to data assimilation schemes, such as optimal interpolation \cite{maeda2015}. Concurrently, sensor placement optimization has become an active research area, with Fujita et al. \cite{fujita2024} combining sensor placement and proper orthogonal decomposition for pseudo-super-resolution, and Wang et al. \cite{wang2020} applying empirical orthogonal function (EOF) analysis to optimally deploy gauges around Crete Island for maximal early warning capability. Notably, these methods are rooted primarily in statistical or physics-based frameworks and have rarely incorporated machine learning (ML) architectures for wavefield reconstruction.

Beyond forecasting arrival times and amplitudes at fixed locations, there is increasing recognition that capturing the spatial dynamics of tsunami waves over wide areas is essential. This not only improves hazard understanding but also supports emergency response, a need highlighted in post-disaster damage assessments using remote sensing and image analysis techniques \cite{koshimura2020}.

Recent ML approaches show promise in forecasting tsunami waveforms at fixed points \cite{liu2021, rim2022tsunami}, but ML solutions for reconstructing full tsunami wavefields from sparse data remain underdeveloped. A related advancement by Archambault et al. \cite{archaumbalt2024} used an attention-based encoder-decoder model to reconstruct sea surface height fields from incomplete satellite altimetry data, leveraging auxiliary sea surface temperature observations and a supervised-to-unsupervised fine-tuning strategy. While their architecture is similar to ours, their problem focuses on spatial interpolation of gridded satellite data, rather than sparse, point-based tsunami sensor networks.

Here, we adapt Senseiver \cite{Santos2023Senseiver}, an attention-based neural network tailored for sparse sensing, capable of reconstructing high-resolution tsunami wavefields from limited tsunameter observations. To focus on the fundamental capabilities of this architecture, we adopt an idealized simulation environment that uses parameterized source models rather than fault-derived displacements. We benchmark Senseiver against the Linear Interpolation with Huygens-Fresnel Principle (LIHFP) method \cite{wang2019}, evaluating accuracy and virtual waveform generation at unobserved locations. Our experiments emulate an operational-like workflow using realistic DART buoy configurations and test generalization to previously unseen epicenters. In addition to generating remarkable full-field reconstructions, results show that Senseiver reduces average errors in wave height, arrival time, and maximum amplitude by approximately a factor of two compared to LIHFP---demonstrating the promise of attention-based ML tools for advancing sensor-informed tsunami forecasting and hazard assessment.

\section{Generating Full-Field Training Data}
\label{sec:pdemodel}

Tsunamis can be effectively modeled using the shallow water equations (SWEs), which describe the dynamics of a fluid layer via a system of partial differential equations for the depth-averaged horizontal velocity $u = u(\mathbf{x},t)$ and fluid thickness $h = h(\mathbf{x},t)$:
\begin{align}
\label{eqn:momentum}
\frac{\partial u}{\partial t} + (u \cdot \nabla)u + fu^{\perp} &= -(1 - \beta) g \nabla (h + z_b) + c_d \frac{|u|}{h}u + \nu_{4}^{u} \nabla^{4} u,
\\[2ex]
\label{eqn:continuity}
\frac{\partial h}{\partial t} + \nabla \cdot (u h) &= 0.
\end{align}
Here, $z_{b} = z(\mathbf{x})$ denotes the bathymetric height at the base of the fluid domain, $g = 9.80665 \mathrm{m}\, \mathrm{s}^{-2}$ is gravitational acceleration, and $f = 2 \Omega \sin \phi$ is the Coriolis force, with $\Omega = 7.292 \times 10^{-5} \mathrm{rad}\, \mathrm{s}^{-1}$ and $\phi$ the latitude. $u^{\perp} = \mathbf{k} \times u$ represents tangential velocity due to the Coriolis effect, where $\mathbf{k}$ is the local vertical. The reduced gravity correction $\beta = 0.015$ accounts for self-attraction and loading effects—that is, changes in sea surface height and ocean dynamics resulting from the gravitational pull of redistributed water mass and the elastic deformation of Earth's crust under that load \cite{inazu2013simulation,2022SAL}. The coefficients $c_d$ and $\nu_{4}$ denote drag and viscous scalings, respectively.

Our dataset consists of time integrations of these equations, with the initial ocean surface displacement parameterized by the epicenter location $\mathbf{x_0}$:
\[
h_0(\mathbf{x};\mathbf{x_0}) = -z_b + 5 \exp{\left(-(250\|\mathbf{x}-\mathbf{x_0}\|^2)^4\right)},
\]
where $\|.\|$ denotes the Euclidean norm. This initial condition produces a localized, flat-topped solitary wave, serving as a simplified yet physically reasonable proxy for tsunami generation due to seafloor deformation. This approach avoids the complexity of explicitly modeling earthquake fault slip---a technically demanding problem and an active area of research~\cite{nhess2024}. While more detailed source models are possible (e.g., elastic dislocation or finite fault modeling), adopting a parameterization based solely on epicenter location enables simplicity of experiment design and subsequent analysis. The 5-meter displacement represents a significant yet plausible event, consistent with historically observed tsunami amplitudes. For details on the numerical methods, including calibration and validation of the PDE model using historical DART data, see Texts S1–S2 and Figures S1–S2 in the Supporting Information.

Importantly, this work is structured as a proof of concept for the Senseiver sparse sensing framework. Our primary goal is to evaluate whether Senseiver can reconstruct high-resolution tsunami wavefields from sparse observations, using the deep-ocean DART system as a realistic baseline. DART buoys were selected specifically because they span broad offshore regions, enabling us to assess Senseiver’s capacity to infer wavefields over large spatial domains—potentially enhancing direct forecasting and improving data assimilation schemes through virtual waveform augmentation, as demonstrated in~\cite{wang2019}.

\begin{figure}[h!]%
\centering
\includegraphics[width=1.0\textwidth]{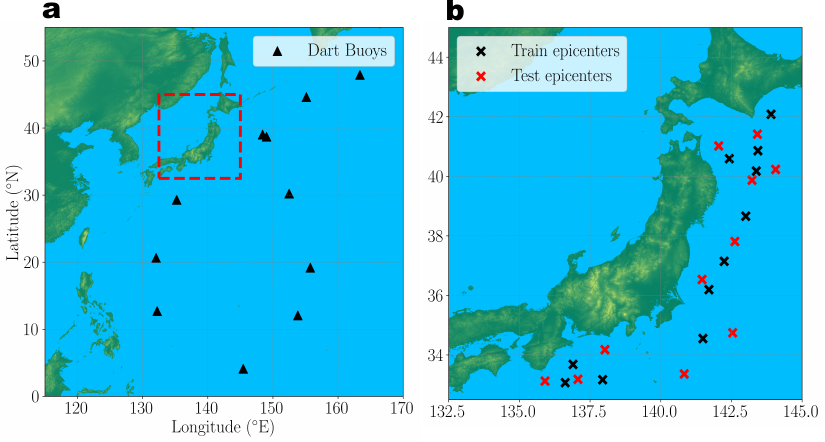}%
\caption{(a) Coverage of DART Buoys, denoted by black triangles. (b) Training (black) and test (red) epicenters used for synthetic tsunami generation.}
\label{fig:train_unseeen_epis}
\end{figure}

 To construct our training and test datasets, we queried the USGS earthquake catalog~\cite{usgs_earthquake_map}, filtering for events that (a) had magnitudes $\geq$ 7.5, (b) listed ``Japan'' in the location descriptor, and (c) occurred in water depths greater than 1000 meters. The depth constraint was adopted to avoid simulation artifacts from initial conditions placed too close to the coast, which can introduce artifacts on coarser meshes. This also ensures the events used are in operationally relevant offshore regions where DART buoys can realistically contribute data. From an initial pool of 92 qualifying events meeting our criteria, we randomly selected 11 epicenters for training. Test epicenters were drawn from the same pool, ensuring each lies within 20 to 100 miles of a training source, to emulate scenarios where events are seismologically related but not spatially redundant. This strategy achieves a balance between data diversity and realistic generalization: the model is exposed during training to a broad distribution of real-world offshore earthquakes, while inference is performed on novel but physically consistent examples. All selected epicenters (see Figure~\ref{fig:train_unseeen_epis}) are located between $136^\circ$–$145^\circ$ longitude and $33^\circ$–$43^\circ$ latitude. The test events lie 20–100 miles from the nearest training epicenter, emulating unseen but seismologically relevant cases. A full pairwise distance matrix is provided in Table~S1.

We acknowledge that for typical operational tsunami forecasting near Japan—especially for events generated close to the coast—DART buoy data alone may not be sufficient due to short arrival times. In practice, such forecasts often rely on coastal instrumentation and denser observation networks. Nevertheless, our focus is not on coastal arrival prediction in isolation, but on evaluating the viability of Senseiver in general settings where sparse observations are the primary inputs. We believe Senseiver’s demonstrated success across a realistic DART configuration strongly suggests it would also perform well in other settings, including denser local arrays and finer model meshes.

Each simulation frame consists of 163,842 unstructured spatial points (longitude, latitude, wave height). To reduce computational load, we subsample frames by a factor of two, resulting in $81,921$ pixels per time step. Simulations span 4 hours, computed at 50-second intervals (289 time steps), yielding a dataset of size $(3179, 81921)$ for each of the training and test sets. Temporal resolution is chosen to balance signal fidelity with computational cost and supports second-order finite difference diagnostics in model validation (see Supporting Information, Text S3).

\section{Methods}
\subsection{The Senseiver}
\label{sec:senseiver}

The sparse sensing problem for tsunami waves can be formulated as follows: Given a discrete set of ocean state observations $s = \{s_{1}, s_{2}, \dots, s_{N}\}$ collected at locations $\mathbf{x}^{s} = \{\mathbf{x}^{s}_{1}, \mathbf{x}^{s}_{2}, \dots, \mathbf{x}^{s}_{N}\}$, the objective is to reconstruct the state $\hat{s} = \{\hat{s}_{1}, \hat{s}_{2}, \dots, \hat{s}_{M}\}$ at a set of query points $\mathbf{x}^{q} = \{\mathbf{x}^{q}_{1}, \mathbf{x}^{q}_{2}, \dots, \mathbf{x}^{q}_{M}\}$ where typically $M \gg N$. In our work, the input state $s(\mathbf{x}^{s})$ corresponds to ocean surface heights measured at DART buoy locations, while the query state $\hat{s}(\mathbf{x}^{q})$ represents the ocean surface height throughout the global ocean domain. Since the DART network consists of only a few dozen buoys, yet even coarse global coverage requires hundreds of thousands of points, this reconstruction task is extremely sparse.

To address this challenge, we employ the Senseiver \cite{santos2023development}, an attention-based neural network specifically designed for sparse-to-dense reconstruction tasks. The Senseiver utilizes a multi-level encoder-decoder architecture that leverages the strengths of attention mechanisms and data compression. Key advantages of the attention operation for sparse sensing include: (1) treating positional information as a feature, (2) being agnostic to grid structure, thus accommodating arbitrary meshes, and (3) enabling immediate long-range spatial associations-unlike CNNs, which require deep layers to integrate distant features. However, attention operations scale quadratically with input size. Senseiver overcomes this bottleneck by encoding the input into compact latent arrays, where most attention computations are performed.

The Senseiver workflow is as follows:
\begin{enumerate}
  \item A positional encoder $P_E$ maps observation locations to spatial encoding vectors $\mathbf{a}^s$.
  \item An attention-based encoder $E$ transforms the observation-value/location pairs $(s_i, \mathbf{a}^s_i)$ into a latent matrix $\mathbf{Z}$.
  \item The positional encoder $P_E$ also maps query locations to spatial encodings $\mathbf{a}^q$.
  \item An attention-based decoder $D$ reconstructs outputs at the encoded query locations $\mathbf{a}^q$.
\end{enumerate}

The positional encoder $P_E$ implements a trigonometric encoding of spatial coordinates, while the encoder and decoder ($E$, $D$) incorporate trainable multi-layer perceptrons within their attention blocks. For a comprehensive description of the Senseiver architecture, see \cite{santos2023development}. The process can be summarized as:
\begin{align}
\mathbf{a}^{s} &= P_{E}(\mathbf{x}^{s})\,, \\[2ex]
\mathbf{Z} &= E(s, \mathbf{a}^{s})\,, \\[2ex]
\mathbf{a}^{q} &= P_{E}(\mathbf{x}^{q})\,, \\[2ex]
\hat{s}(\mathbf{x}^{q}) &= D(\mathbf{Z}, \mathbf{a}^{q})\,.    
\end{align}

\begin{figure}[h!]%
\centering
\includegraphics[width=1.05\textwidth]{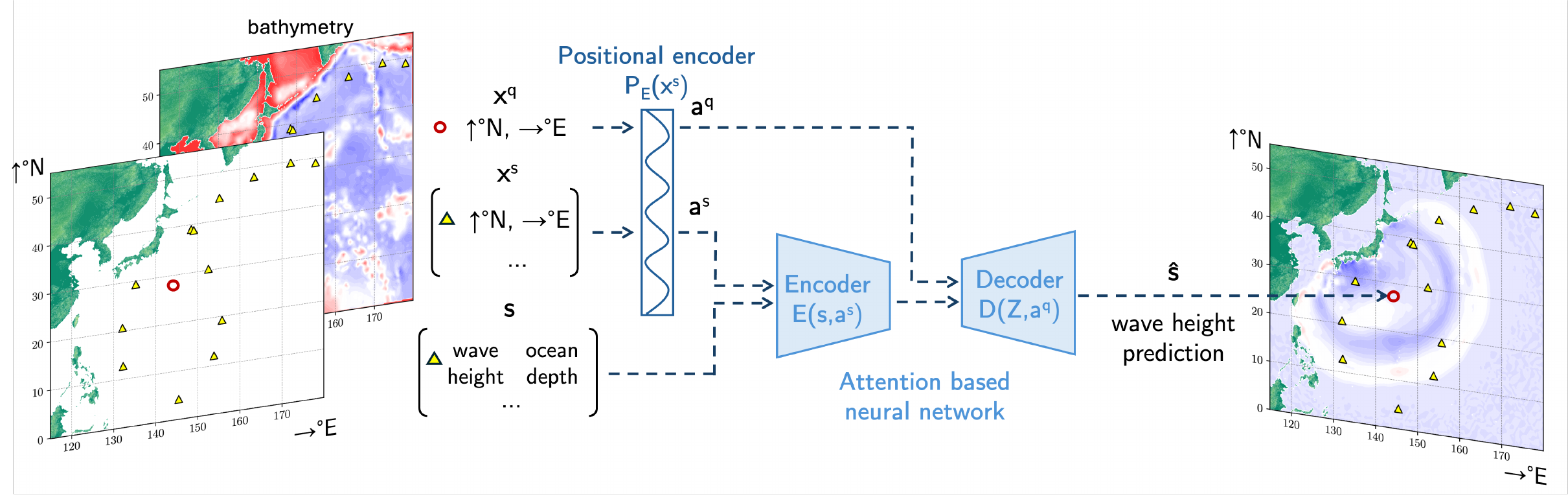}%
\caption{Overview of the Senseiver architecture for tsunami reconstruction. The sensor values along with their encoded positions are processed by the encoder into a latent array $Z$ of fixed size. The encoded query locations are processed by the attention-based decoder, which outputs an estimate of wave height at the query location(s).}
\label{fig:sens_workflow}
\end{figure}

This flexible architecture enables modeling on unstructured data and allows for custom coordinate choices in the positional encoder. In our application, we augment latitude and longitude with ocean bathymetry encodings. Land pixels are masked out, ensuring training occurs only on ocean data. A schematic of the Senseiver workflow is presented in Figure~\ref{fig:sens_workflow}.

Model weights are optimized by minimizing the mean squared error:
\begin{gather}
  \label{eqn:training_loss}
  \mathcal{L} = \sum \Big( 
    s(\mathbf{x}^{q}) - 
    \hat{s}(\mathbf{x}^{q})\Big)^2\,, \\[2ex]
  \text{where}\quad \hat{s}(\mathbf{x}^{q}) = D\left(
      E\left(s(\mathbf{x}^{q}), P_{E}(\mathbf{x}^{s})\right), P_E(\mathbf{x}^{q})
    \right),
\end{gather}
using the Adam optimizer \cite{kingma2014adam}. To prevent overfitting, we train on only $80\%$ of the available training frames. The number and ordering of query points, as well as the data frames used in training, are treated as hyperparameters, with further details provided in Text S4 of the Supporting Information.

\subsection{Experiments}

The primary error metric used to evaluate reconstruction quality is the mean absolute error (MAE), normalized by the absolute maximum wave height in each frame. To ensure this metric captures meaningful differences, we compute the MAE only over pixels where the true wave height exceeds $1 \times 10^{-4}$. This threshold is necessary because most of the ocean domain has zero wave height; including these pixels would artificially deflate the error. For each reconstruction, we calculate the absolute difference between the true and reconstructed wave height fields, divided by the maximum absolute true wave height for that frame:
\begin{equation}
\label{eqn:recon_loss}
\text{Error}(x,y,t) = \frac{|h(x,y,t)-\hat{h}(x,y,t)|}{\text{max}(|h(x,y,t)|)}.
\end{equation}

While this metric provides a standard quantitative assessment typical in machine learning, we also sought to evaluate our approach in a more operationally relevant context. In practice, tsunami forecasting systems such as the tsunami data assimilation method (TDAM) \cite{maeda2015} focus on predicting key features of tsunami waveforms—specifically, arrival times and maximum amplitudes—at fixed observation points. TDAM performs well with dense observation networks, but its accuracy diminishes with sparse sensor coverage. To address this, the Linear Interpolation with Huygens-Fresnel Principle (LIHFP) \cite{wang2019} was developed to augment sparse networks, estimating virtual waveforms at unsensed locations by interpolating between real sensor data while accounting for arrival times, distances, and bathymetry.

To bridge the gap between standard ML evaluation and practical tsunami forecasting, we compare our model’s reconstructions not only in terms of full-field MAE, but also by generating six virtual waveforms at strategically chosen locations and benchmarking them against those produced by the LIHFP method. This dual evaluation strategy allows us to assess both the general reconstruction capability of our model and its effectiveness in a real-world forecasting workflow.

\section{Results}
Averaging the reconstruction loss (Eq.~\ref{eqn:recon_loss}) over all 289 frames from each of the eleven test epicenters yields a global experiment error of $5.00 \times 10^{-2}$. Notably, much of this error arises from the early time frames, when sensors have yet to detect significant tsunami signals. This limitation is inherent to any assimilation method that relies solely on tsunameter data. On average, the reconstruction error remains below $1 \times 10^{-1}$ after 32.9 minutes, with a median threshold time of 38.3 minutes. To assess the physical consistency of our model’s reconstructions, we evaluated the continuity equation residual (Eq.~\ref{eqn:continuity}), yielding a global average error of $4.10 \times 10^{-2}$ across all times and epicenters. Details of the diagnostic calculation and frame-by-frame residual plots are provided in the Supporting Information (Text S3, Figures S15). The key results from the reconstruction experiments for all eight simulations are summarized in the first seven columns of Table~\ref{table:table}.

\begin{figure}[h!] 
\centering{
\includegraphics[width=.90\textwidth]{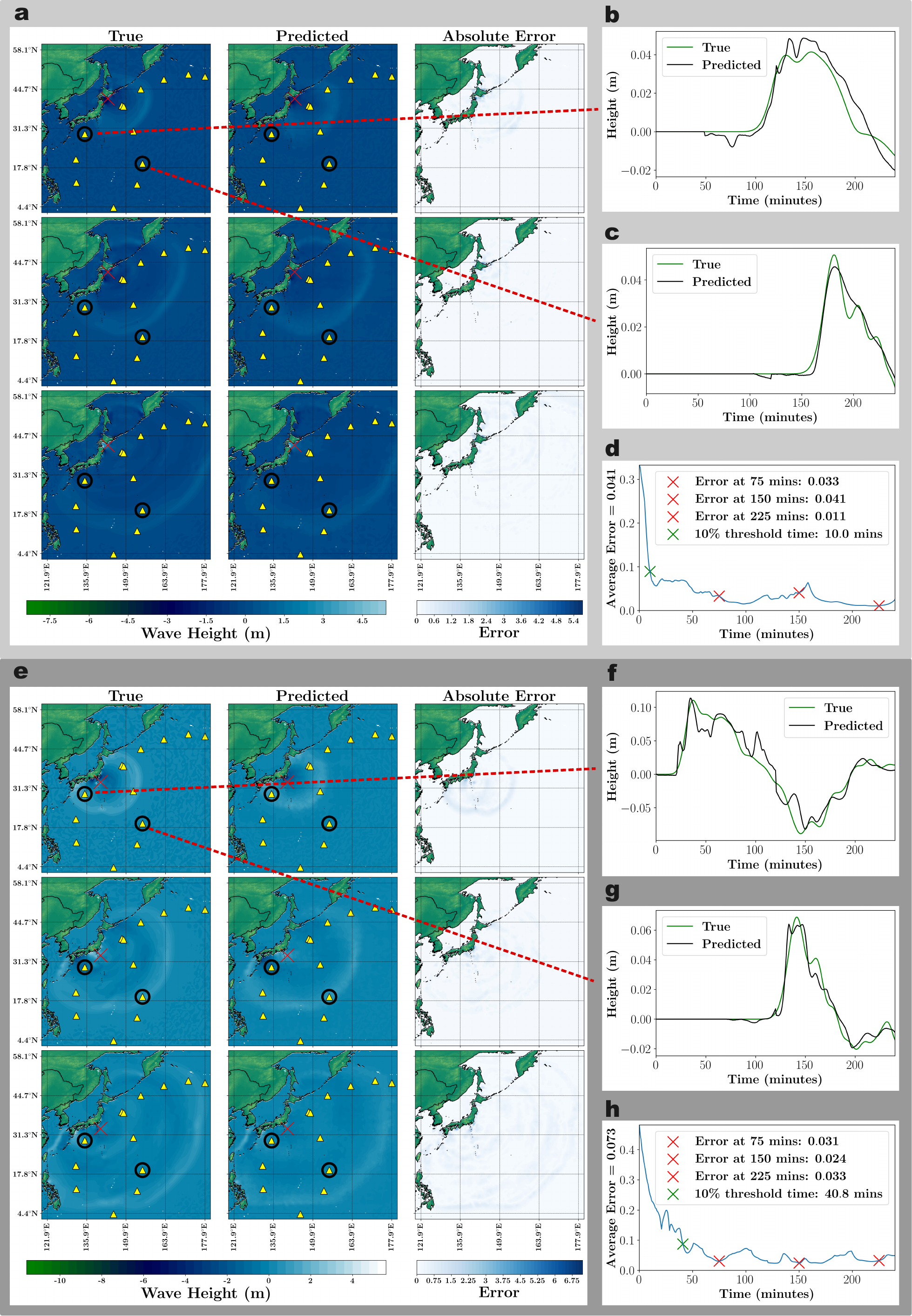}%
}
\caption{(a) Full reconstruction at $75$, $150$, and $225$ minutes for test epicenter $(143.4^\circ E, 41.4^\circ N)$. The epicenter location is indicated by an X, and yellow triangles notate the sensors. (b)-(c) Waveform reconstructions at fixed DART buoys with longitude-latitude coordinate $(135.2^\circ E,29.3^\circ N)$ and $(155.7^\circ E,19.2^\circ N)$, respectively from top to bottom. (d) Per-pixel average reconstruction error as a function of time for test epicenter $(143.4^\circ E, 41.4^\circ N)$. (e)-(h) Equivalent figures for test epicenter $(140.8^\circ E, 33.4^\circ N)$}
\label{fig:fig3}
\end{figure}

Figure~\ref{fig:fig3} presents representative reconstructions for two test simulations, with epicenters at $(143.4^\circ E, 41.4^\circ N)$ and $(140.8^\circ E, 33.4^\circ N)$, shown at 75, 150, and 225 minutes. For each case, we include example waveform reconstructions at fixed DART buoy locations and the per-pixel average reconstruction error over time. These two epicenters were chosen to illustrate model performance at different distances from the training set: the first is the median-distance epicenter (38.4 miles from the nearest training example), while the second is the most distant (90.6 miles). Together, they demonstrate reconstruction accuracy for both typical and challenging test cases. Reconstructions, error plots, and time series for all other test epicenters are provided in Figures S3–S14 of the supporting information.

Visual inspection shows that the Senseiver effectively captures key features of the tsunami wavefield, including the spatial support, number of wave periods, and amplitude. The error plots reveal that reconstruction error decreases rapidly within the first $\approx$ 50 minutes, highlighting the particular challenge of early-time reconstruction when little sensor information is available. For the median-distance epicenter, the error drops below $1 \times 10^{-1}$ after just 10 minutes; for the most distant epicenter, this threshold is reached at 40.8 minutes. While resolving the earliest time frames is not the focus of this work, we suggest alternative datasets that could better inform the Senseiver at early time frames in the conclusion.

To ensure consistency with the experimental setup of \cite{wang2019} and enable direct comparison, we focused our evaluation on virtual waveform reconstruction at midpoints between selected DART buoys. Although our experiments center on these midpoints, the methodology is readily extensible to arbitrary virtual sensor locations. In fact, analyzing Senseiver performance as the virtual observation point moves farther from the training data may provide valuable insight into the method’s effective range—potentially revealing where LIHFP waveforms surpass Senseiver reconstructions in accuracy. For this study, we identified six midpoints among DART buoys that registered significant waveforms in our simulations and compared the Senseiver’s reconstructions at these locations to those produced by LIHFP.

\begin{figure}[h!] 
\centering{
\includegraphics[width=.90\textwidth]{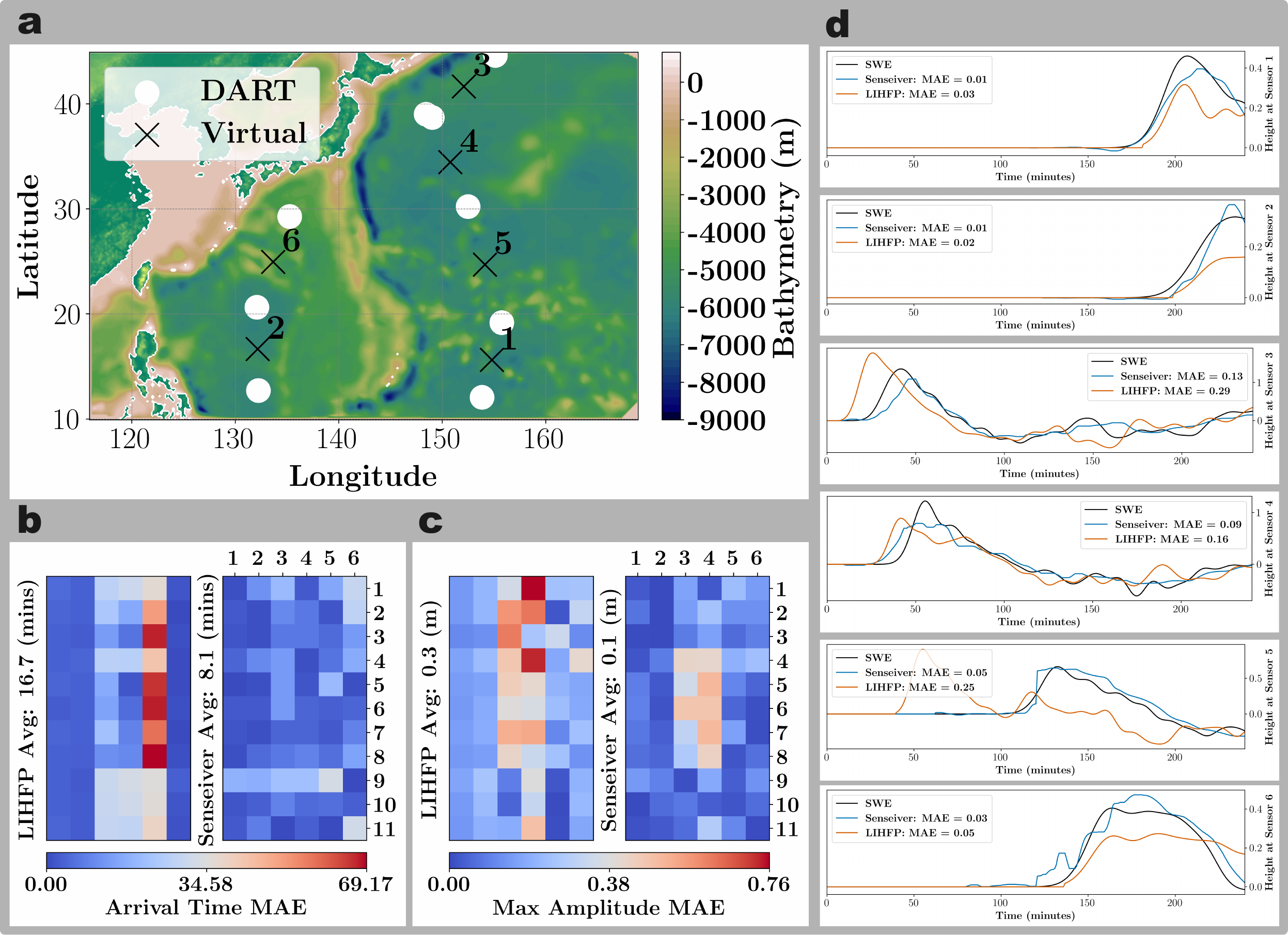}%
}
\caption{(a) Observation network augmented with virtual sensors with locations derived as midpoints between DART buoys. (b)-(c). Mean absolute error in arrival time and maximum amplitude, respectively. Each color map (6 columns $\times$ 11 rows) represents all combinations of test epicenter and virtual observation pairs.  (d) Senseiver and LIHFP-based virtual waveform reconstructions at test epicenter $(143.4^\circ E, 41.4^\circ N)$.}
\label{fig:fig4}
\end{figure}

Across the majority of test epicenters, times, and virtual observation sites, the Senseiver outperformed LIHFP, achieving a mean absolute error (MAE) in wave height of $7.02 \times 10^{-2}$ meters, compared to $18.2 \times 10^{-2}$ meters for LIHFP. The Senseiver also demonstrated substantial improvements in estimating both arrival times and maximum amplitudes at nearly all virtual observation points. Specifically, the mean absolute error for arrival time, averaged across all eleven simulations, was $8.12$ minutes for the Senseiver and $16.68$ minutes for LIHFP. The only exception occurred at the epicenter located at $(135.9^\circ E, 33.1^\circ N)$, where the Senseiver’s arrival time MAE was $18.611$ minutes, slightly higher than the $17.778$ minutes observed for LIHFP. For maximum amplitude, the Senseiver achieved an MAE of $13.1 \times 10^{-2}$ meters, outperforming LIHFP’s MAE of $25.9 \times 10^{-2}$ meters across all cases. On average, the percent change in arrival time MAE for the Senseiver was $-51.9\%$, and for maximum amplitude, it was $-49.9\%$, indicating a substantial reduction in error compared to LIHFP. These results are summarized in Figure~\ref{fig:fig4} (b) and (c), which display arrival time and maximum amplitude errors for both methods across all 48 test epicenter–virtual observation point pairings. Key numerical results for all eleven simulations can be found in the last six columns of Table~\ref{table:table}.

Figure~\ref{fig:fig4} (d) provides a detailed summary for the epicenter at $(143.4^\circ E, 41.4^\circ N)$ (the median epicenter). For this case, the Senseiver achieved a substantially lower MAE across the six virtual waveforms: $5.5 \times 10^{-2}$ meters, compared to $13.3 \times 10^{-2}$ meters for LIHFP. Arrival time errors were similarly reduced, with the Senseiver yielding an MAE of $7.22$ minutes versus $16.39$ minutes for LIHFP. For maximum amplitude, the Senseiver achieved an MAE of $14.9 \times 10^{-2}$ meters, again outperforming LIHFP’s $23.8 \times 10^{-2}$ meters. Virtual waveform comparisons for all remaining test epicenters are provided in Figures S16–S24 of the supporting information.

We conclude this section by noting the lack of temporal smoothness observed in the waveforms generated by the Senseiver. To mitigate this issue, we applied median filtering—a non-linear technique that replaces each point in a signal with the median value within a local window, effectively reducing noise and outliers while preserving key features—in our virtual waveform experiments. This post-processing step substantially improved arrival time accuracy: for arrival times defined by a 3 cm threshold, repeated median filtering reduced the raw Senseiver error from $9.05$ minutes to $5.35$ minutes. However, this improvement came at the cost of a decrease in maximum amplitude accuracy. For the plots in Figure~4 and the statistics collected in Table ~\ref{table:table}, Senseiver signals were processed using a single median filter with a kernel length of 13. Analogous statistics are provided for the raw Senseiver waveforms as well as heavily filtered senseiver waveforms in tables S2-S3 of the supporting information. Addressing temporal smoothness directly within the machine learning framework—for example, by incorporating temporal encodings, conditioning on previous states, or introducing regularization terms into the loss function—remains an important direction for future research.

\begin{table}[ht]
\centering
{\footnotesize
\begin{tabular}{|P{1.5cm}|P{1.5cm}|P{1.5cm}|P{1.5cm}|P{1.5cm}|P{1.5cm}|P{1.5cm}|}
\hline
\textbf{Epicenter Number} & \textbf{Longitude} & \textbf{Latitude} & \textbf{Closest Distance from Training Set} & \textbf{Mean Recon Error} & \textbf{Mean Physics Error} & \textbf{Trigger Time (mins)} \\
\hline
1 & 140.827 & 33.362 & 90.551 & 0.073 & 0.047 & 40.8 \\
\hline
2 & 141.459 & 36.534 & 27.105 & 0.045 & 0.043 & 38.3 \\
\hline
3 & 142.050& 41.019& 35.398 & 0.056 & 0.045 & 46.7 \\
\hline
4 & 142.542 & 34.745 & 61.320 & 0.062 & 0.040 & 43.3 \\
\hline
5 & 143.228 & 39.869 & 22.648 & 0.036 & 0.036 & 14.2 \\
\hline
6 & 144.060 & 40.232 & 36.197 & 0.042 & 0.039 & 39.2 \\
\hline
7 & 142.619 & 37.812 & 50.405 & 0.040 & 0.035 & 24.2 \\
\hline
8 & 143.416 & 41.415 & 38.360 & 0.041 & 0.045 & 10.0 \\
\hline
9 & 135.905 & 33.123 & 41.433 & 0.061 & 0.043 & 70.8 \\
\hline
10 & 137.071 & 33.184 & 27.370 & 0.035 & 0.036 & 11.7 \\
\hline
11& 138.025 & 34.175 & 69.802 & 0.057 & 0.041 & 22.5 \\
\hline
\multicolumn{7}{|c|}{
  \begin{tabular}{P{1.5cm}|P{1.5cm}|P{1.5cm}|P{1.5cm}|P{1.5cm}|P{1.5cm}}
    \hline
    \textbf{Senseiver Arrival Time MAE (mins)} & \textbf{LIHFP Arrival Time MAE (mins)} & \textbf{Senseiver Max Amp MAE (m)} & \textbf{LIHFP Max Amp MAE (m)} & \textbf{Senseiver Wave Height MAE (m)} & \textbf{LIHFP Wave Height MAE (m)} \\
    \hline
    13.333 & 17.361 & 0.100 & 0.326 & 0.120 & 0.246 \\
    \hline
     8.333 & 16.458 & 0.121 & 0.316 & 0.064 & 0.211\\
    \hline
     5.972 & 15.833 & 0.067 & 0.261 & 0.054 & 0.117 \\
    \hline
     8.611 & 17.361 & 0.229 & 0.363 & 0.103 & 0.268\\
    \hline
     6.111 & 15.417 & 0.186 & 0.258 & 0.054 & 0.160 \\
    \hline
     5.000 & 15.903 & 0.185 & 0.248 & 0.069 & 0.167 \\
    \hline
     3.750 & 16.181 & 0.182 & 0.291 & 0.073 & 0.201 \\
    \hline
     7.222 & 16.389 & 0.149 & 0.238 & 0.055 & 0.133\\
    \hline
     18.611 & 17.778 & 0.100 & 0.187 & 0.087 & 0.165 \\
    \hline
     4.722 & 17.431 & 0.043 & 0.161 & 0.047 & 0.173 \\
    \hline
     7.639 & 17.361 & 0.082 & 0.198 & 0.068 & 0.164 \\
    \hline
  \end{tabular}
} \\
\hline
\end{tabular}
}
\caption{Overview of Senseiver evaluation metrics for all test epicenters. Columns 4-7 correspond to reconstruction experiments, and columns 8-13 correspond to LIHFP comparisons. MAE refers to mean absolute error, and Trigger Time refers to the time at which no reconstruction exceeds $1\times 10^{-1}$.}
\label{table:table}
\end{table}

\section{Conclusion}
We have presented Senseiver, an attention-based neural network capable of reconstructing high-resolution tsunami wavefields from sparse observation networks. Our results demonstrate substantial improvements over the LIHFP baseline, particularly in generating accurate virtual waveforms at unobserved locations. To our knowledge, this is the first ML-based solution to address the sparse sensing problem for tsunami forecasting at high spatial resolution.

This work serves as a proof of concept, with several important limitations. Our experiments are restricted to a limited oceanic region and idealized initial conditions. Future research should extend these methods to broader domains, diversify training data, and investigate how regional scale interacts with model complexity. Incorporating additional features and regularization strategies may further improve the temporal smoothness of Senseiver-generated waveforms.

A key challenge remains early-stage reconstruction, before tsunameter data are available. Future work could explore integrating alternative data sources such as GNSS \cite{rim2022tsunami}, HF radar \cite{wang2023}, or distributed acoustic sensing \cite{xiao2024} to provide earlier or more robust wavefield estimates. Additionally, adapting Senseiver to optimize sensor placement during training \cite{marcato2023reconstruction} could inform the design of observation networks, particularly in resource-limited regions.

By bridging advances in machine learning with operational tsunami sensing, Senseiver offers a promising step toward more accurate and timely tsunami early warning systems.

\section*{Open Research Section}
The tsunami simulation data used in this study are available from Zenodo \url{https://zenodo.org/records/15478821} \cite{McDugald2025TsunamiSims}. The Senseiver codebase is available from GitHub \url{https://github.com/OrchardLANL/Senseiver} \cite{Santos2023Senseiver}, with tsunami-specific code provided in the ``tsunami'' directory.  Formal citations for both the dataset and software are included in the reference list.

\acknowledgments
This project was partially funded by the Artimis LDRD program at the Los Alamos National Laboratory. The Los Alamos Unlimited Release number for this work is LA-UR-24-32151. Andrew Arnold (University of Arizona PhD Student in Applied Mathematics) assisted in analyzing the virtual waveform experiments.

\clearpage
\bibliography{refs}

%
%


%
%
%
%
%

\end{document}